\documentclass[conference]{IEEEtran}
\IEEEoverridecommandlockouts

\usepackage{hyperref}
\usepackage[numbers,sort&compress]{natbib}

\usepackage[printonlyused]{acronym}

\usepackage{siunitx}
\usepackage[all]{nowidow}
\usepackage{trimclip}
\usepackage{tikz}
\usetikzlibrary{arrows}
\usetikzlibrary{arrows.meta}
\usepackage{makecell}
\usetikzlibrary{positioning}
\usetikzlibrary{decorations.pathreplacing,calligraphy}
\usepackage[normalem]{ulem}

\usepackage{lipsum}

\usepackage{hhline}

\usepackage{xspace} 

\usepackage{epstopdf}

\usepackage{import}

\usepackage[table]{xcolor}
\usepackage{booktabs}
\usepackage{array}
\usepackage{longtable}
\newcolumntype{L}{>{\raggedright\arraybackslash}p{0.13\textwidth}}
\newcolumntype{C}{>{\centering\arraybackslash}p{0.10\textwidth}}
\newcolumntype{Y}{>{\centering\arraybackslash}m{1.05cm}}
\newcolumntype{T}{>{\centering\arraybackslash}m{1.5cm}}
\newcolumntype{O}{>{\centering\arraybackslash}m{1.75cm}}
\newcolumntype{M}{>{\centering\arraybackslash}p{0.065\textwidth}}

\usepackage[outline]{contour}
\usepackage{xcolor}

\usepackage{tabularx}
\usepackage{multirow, multicol}

\newlength{\Oldarrayrulewidth}

\newcolumntype{?}[1]{!{\vrule width #1}}

\usepackage{amssymb,amsfonts,amsmath,amscd}

\usepackage{pifont}
\usepackage{bm}

\usepackage{balance}
\usepackage[T1]{fontenc}

\usepackage{cancel}

\newcommand{\bbm}{\begin{bmatrix}}
\newcommand{\ebm}{\end{bmatrix}}

\newcommand{\ignore}[1]{}

\newcommand{\bma}[1]{\left[\begin{array}{#1}}
\newcommand{\ema}{\end{array}\right]}

\DeclareMathAlphabet{\mbf}{OT1}{ptm}{b}{n}

\def\fdotb{{\raisebox{-0.6ex}{ \kern0.2ex\raisebox{0.8ex}{\tiny $\hspace*{-1ex}\circ$}}}}
\def\fddotb{{\raisebox{-0.6ex}{ \kern0.2ex\raisebox{0.8ex}{\tiny $\hspace*{-1ex}\circ\circ$}}}}

\newcommand{\trans}{{\ensuremath{\mathsf{T}}}} 
\newcommand{\utimes}{ {\raisebox{-0.6ex}{ \kern-1.0ex\raisebox{0.6ex}{ \small $\mathsf{v}$}}} } %
\newcommand{\beq}{\begin{equation}}
\newcommand{\eeq}{\end{equation}}
\newcommand{\bdis}{\begin{displaymath}}
\newcommand{\edis}{\end{displaymath}}
\newcommand{\beqarray}{\begin{eqnarray}}
\newcommand{\eeqarray}{\end{eqnarray}}
\newcommand{\beqarraynn}{\begin{eqnarray*}}
\newcommand{\eeqarraynn}{\end{eqnarray*}}

\DeclareMathAlphabet{\mbf}{OT1}{ptm}{b}{n}

\DeclareMathOperator*{\argmin}{arg\,min}

\acrodef{FMCW}[FMCW]{frequency modulated continuous wave}
\acrodef{DRO}[DRO]{direct radar odometry}
\acrodef{DRL}[DRL]{direct radar localization}
\acrodef{ICP}[ICP]{iterative closest point}
\acrodef{GNSS}[GNSS]{global navigation satellite system}
\acrodef{RMS}[RMS]{root mean squared}
\acrodef{RMSE}[RMSE]{root mean squared error}
\acrodef{INS}[INS]{inertial navigation system}
\acrodef{SOTA}[SOTA]{state-of-the-art}
\acrodef{MPC}[MPC]{model-predictive controller}
\acrodef{IMU}[IMU]{inertial measurement unit}

\acrodef{tr}[T\&R]{teach and repeat}
\acrodef{rtr}[RT\&R]{radar teach and repeat}
\acrodef{drtr}[DRT\&R]{direct radar teach and repeat}
\acrodef{ltr}[LT\&R]{lidar teach and repeat}
\acrodef{sota}[SOTA]{state-of-the-art}

\begin{document}

\title{DRT\&R: Direct Radar Teach \& Repeat\\
\thanks{$^*$ Authors contributed equally. \\
$^1$ Robotics Institute,
University of Toronto, Canada \\
$^2$ Mobile Robotics Lab, ETH Z\"{u}rich, Switzerland \\
Corresponding: Alexander Krawciw, alec.krawciw@robotics.utias.utoronto.ca}
}

\author{\IEEEauthorblockN{Alexander Krawciw$^{*1}$, Daniil Lisus$^{*1}$, Cedric Le Gentil$^2$, Timothy D. Barfoot$^1$}
}


\maketitle

\begin{abstract}
Radar-based navigation is appealing for its robustness to adverse conditions involving airborne particles, such as precipitation, dust, fog, and smoke, that can cause lidar-based systems to fail.
Recently, direct methods that retain and use the entire radar scan rather than sparse points have improved on-road global localization performance.
However, they have yet to be deployed in off-road environments or in closed-loop systems.
Additionally, even direct global maps may lose information: their global nature leads to a smoothing out of viewpoint-dependent radar artifacts, which can provide pose information when mapping and localization occur along similar trajectories.
This paper introduces Direct Radar Teach \& Repeat (DRT\&R): a direct spinning radar-based navigation stack that maximizes the amount of retained information by combining direct radar processing with local mapping.
DRT\&R yields state-of-the-art (SOTA) localization performance in both on-road and off-road environments.
Using $344\;\si{\km}$ of on-road data and $20\;\si{\km}$ of off-road data, DRT\&R is able to localize to within $4\; \si{\cm}$ in most on-road and off-road conditions, and $12\; \si{\cm}$ in geometrically degenerate and sparse environments.
DRT\&R is also evaluated autonomously in closed loop with an MPC controller for more than $10\;\si{\km}$ using a Clearpath Warthog off-road vehicle, demonstrating that it runs in real time and achieves SOTA tracking performance for off-road radar navigation.
\end{abstract}



\section{Introduction}

\Ac{FMCW} spinning radar sensors have been gaining popularity in robotics for their 360$^\circ$ viewpoint, weather resilience, and long range \cite{a_new_wave_radar}.
To date, most work has focused on odometry and localization, typically through the extraction of sparse point clouds from range-intensity returns \cite{are_we_ready_for, adolfsson2023cfear, cen2019, hilger2026cfear}.
Although meaningful performance gains have been realized in structured, predominantly planar on-road environments, generalization to unstructured off-road deployment remains limited, with performance degrading to the point of failure in such conditions \cite{kolhe_2026, fomo_dataset}.
In addition, \ac{rtr} \cite{rtr_1}, the only purely radar-based closed-loop work to date, shows degradation in performance in fully off-road scenarios.

\begin{figure}
    \centering
    \includegraphics[width=0.9\linewidth]{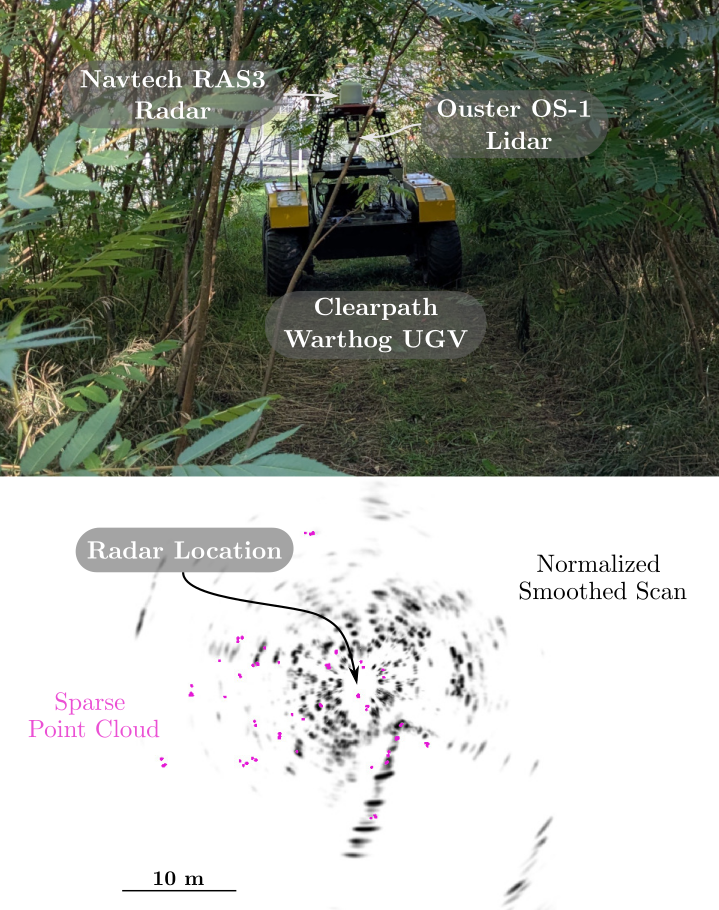}
    \caption{Top: The Clearpath Warthog Uncrewed Ground Vehicle (UGV) as it drives through overhanging brush, highlighting the spinning radar used for navigation and the lidar used for benchmarking. Bottom: An overlay of a sparse radar point cloud over a DRO smoothed scan. Direct methods such as \ac{drtr} retain more scene information than sparse point-based ones, improving localization and path tracking.}
    \label{fig:intro}
    \vspace{-3mm}
\end{figure}

Recently, direct methods operating on the full radar return have shown significant performance gains on-road \cite{legentil2025dro, legentil2026drpogo, lisus2026drba}, motivating their application for off-road.
One factor to consider when using direct methods is that estimation is done with unique radar artifacts present in the measurements.
These artifacts have been traditionally viewed as something to be removed prior to using radar data for state estimation, as they do not correspond to physical objects in the environment.
Point cloud methods are typically tuned or modified to avoid extracting points from artifacts \cite{finer_points, lisus2025pointing}, while direct methods try to smooth out or remove the pixels associated with them \cite{masking_by_moving, legentil2025dro, lisus2026drba}.
However, analysis \cite{legentil2025dro, lisus2026drba} of the Boreas Road Trip (Boreas-RT) dataset \cite{lisus_brrt26} reveals locations containing significant radar artifacts that are consistent between multiple observations.
This suggests that they could be a rich source of localization information when encoded into the reference map.
While encoding pose dependence is challenging in globally aligned maps, a collection of local maps can reduce errors in sensors with state-dependent biases by incorporating this dependence directly~\cite{tr_review_2026}.
\Ac{tr} \cite{VTR} is an appealing framework for spinning radar because it uses local maps connected to a safe, taught path and path-tracking control to guide the robot to repeat autonomously, ensuring that sensor viewpoints are consistent while localizing.

Motivated by advances in direct radar methods and the observation of repeatable radar artifacts, we combine \ac{DRO} \cite{legentil2025dro} and \ac{DRL}~\cite{lisus2026drba} into \acf{drtr} and evaluate the full system in closed-loop off-road on a Clearpath Warthog.
Fig.~\ref{fig:intro} shows the vehicle as it drives in close proximity to foliage.
Comparison of the \ac{DRO} smoothed scan and extracted point clouds in this environment illustrates the additional available information for direct methods over sparse ones.
We evaluate both pure localization performance, particularly against the \ac{SOTA} direct globally consistent mapping method \cite{lisus2026drba}, and closed-loop path tracking against the existing point cloud-based method \cite{rtr_1}.
Since \ac{GNSS} performance is poor in some of the off-road environments, a lidar-based ground-truthing solution is used instead as a reference.
A summary of our contributions is as follows:
\begin{itemize}
    \item \Ac{drtr}: the first direct radar topometric localization and navigation approach.
    \item New \ac{SOTA} radar-based localization results on $344\;\si{\km}$ of on-road data and $20\;\si{\km}$ of off-road data. 
    \item Closed-loop path tracking over $10\;\si{\km}$ of autonomous off-road driving with $6\; \si{\cm}$ \ac{RMS} error, on par with lidar-based methods.
    \item A publicly available $20\;\si{\km}$ off-road spinning radar, $360\si{\degree}$ lidar, and IMU dataset with lidar-based groundtruthing\footnote{Dataset to be released upon acceptance to maintain anonymity.}.
\end{itemize}
\section{Related Work}

Radar-based localization has two main design decisions: intermediary data representation and localization framework architecture.
We present an overview of these choices here and point the reader to \cite{nader2024survey, a_new_wave_radar, venon2022millimeter} for a broader summary of the use of radar in state estimation.

Data representations vary between sparse feature-based methods, such as those using point clouds~\cite{cen2018precise, adolfsson2023cfear} or learned features~\cite{callmer2011radarslamusingvisualfeatures, under_the_radar}, and direct methods, which use the entire intensity return per azimuth and are typically rooted in some form of a cross-correlation between different radar scans~\cite{masking_by_moving, Checchin2009, park2020pharao, legentil2025dro}.
Feature-based approaches have been the default choice for early radar state estimation due to the rich variety in point cloud extraction approaches borrowed from a long history of radar use \cite{finer_points}.
However, sparse representations tend to rely on consistent, strongly reflective objects and general structure in the environment.
For off-road robotics, these assumptions are inherently violated due to foliage, few human-made, highly reflective objects, and a lack of geometric features.
Fig.~\ref{fig:intro} shows a radar scan in a semi-structured environment, where it is challenging to extract reliable points due to the brush.
Indeed, evaluation of \ac{ICP}-based radar odometry and localization methods that work on-road has observed failures in off-road datasets \cite{fomo_dataset, boxan_2026, kolhe_2026}.
Direct methods are a promising alternative, since they do not downsample the scene based on prior assumptions about what kind of returns would make for reliable state estimation.
\Ac{DRO} \cite{legentil2025dro}, explained in more detail in Section~\ref{sec:teach}, showed that a direct odometry method was more robust and accurate than a point cloud one in both on-road and off-road environments.
\Ac{DRO} was extended to pose-graph optimization \cite{legentil2026drpogo} and global localization \cite{lisus2026drba}, where similar robustness and performance was shown on-road, but not studied off-road.

The other design choice is in whether localization is done relative to a global or topometric map.
Global methods typically rely on loop-closure detections \cite{jang2023raplace, jang2025xpress, kim2022scancontextpp} to produce consistent maps across tens of kilometres.
The backbone of these methods is typically pose-graph optimization~\cite{adolfsson2023tbv, holder2019realtime, hong2022radarslam}, which produces a set of globally consistent poses, or full bundle adjustment or SLAM, which produces both globally consistent poses and a consistent map \cite{lisus2026drba, callmer2011radarslamusingvisualfeatures}.
Globally consistent maps are sometimes required for downstream applications; however, they may yield worse localization performance due to their tendency to remove pose-specific biases from the measurements~\cite{tr_review_2026}.
Topometric localization overcomes this issue by keeping track of only locally consistent, but globally interconnected, metric maps.
In this work, we use the \ac{tr} approach, a well-established framework for precise navigation in \ac{GNSS}-denied off-road environments~\cite{VTR}, for topometric spinning radar localization.
Offline topometric radar localization has thus far only used point cloud methods.
\citet{are_we_ready_for} use sparse points extracted from the radar to perform odometry and localization in a topometric approach against both radar and lidar point cloud maps.
\citet{rtr_1} adapt this approach for a closed-loop radar-only \ac{tr} method.
\citet{tro_xiao_2026} adopt an entirely multi-modal approach; they use lidar to create the topometric maps and 4D automotive radar to localize against them during repeats.
\citet{hilger2026cfear} integrate CFEAR odometry \cite{adolfsson2021CFEAR} into a topometric \ac{tr} pipeline which stores maps as points with oriented normals.
\citet{tro_xiao_2026} and \citet{rtr_1} are the only methods to-date that deploy radar \ac{tr} in a closed-loop on a robot, with the latter showing that performance depends on the amount of structure in off-road environments.
Our approach shows that deploying direct methods in a \ac{tr} approach yields significantly improved offline localization and closed-loop tracking performance that does not degrade in unstructured environments.
\section{Method}

\begin{figure*}[t!]
    \centering
    \resizebox{0.85\linewidth}{!}{\input{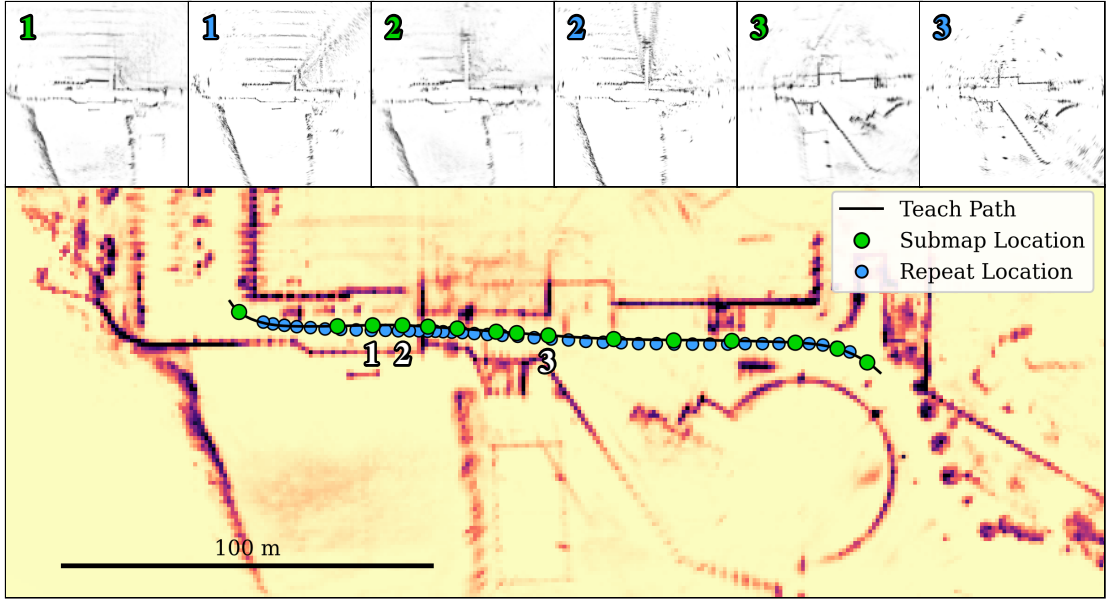}}
    \caption{A globally aligned bird's-eye view map produced by Dr-BA \cite{lisus2026drba} (bottom), overlaid with submap locations (green) and repeat locations (blue) created during the \textit{teach} and \textit{repeat} phases of our DRT\&R approach. Selected submaps and `live' repeat smoothed scans are visualized on top. Submaps enable encoding of perspective-dependent features that are averaged out in a global map, but are useful for localizing: radar artifacts such as multi-path reflections (top of 1) and viewpoint-specific geometry (vertical road in 2, extended fence in 3). These features are re-observed during the repeat runs at similar locations.}
    \label{fig:garage_submapping}
\end{figure*}

\Ac{drtr} adapts the \ac{tr} approach to direct radar methods.
\Ac{tr} consists of two stages: \textit{teach} and \textit{repeat}.
During a teach, the robot is manually piloted through an environment of interest, creating a topometric map as it goes. 
These topometric maps link places of interest topologically along paths that are known to be traversable. 
At regular locations along the path, local maps, or submaps, are created containing data required for an exteroceptive sensor to perform metric localization.
The linking edges store the relative motion estimated by odometry, creating locally consistent paths and submaps for the robot to use. 
A repeat occurs when the robot autonomously navigates within the network of paths created while teaching, using odometry as a fallback and prior for localization to the nearest submap. 
The most recent successful localization and odometry guide the system to switch submaps as it drives.
The main differentiating features of any teach-and-repeat system are the odometry method used to estimate the teach path, the manner in which the submaps are represented, and how the robot is localized within the submaps.
In \Ac{drtr}, odometry, submaps, and localization are all rooted in dense \ac{FMCW} radar scans.

\subsection{Radar Formulation}
Consider a spinning radar scan with $N$ azimuths and $M$ range bins.
Each row of the image corresponds to a measured azimuth angle, $\alpha_n$, at measurement time $t_n$.
At every range bin $\hat{r}_m$ there is an intensity measurement $\phi_{nm}$.
The radar data is the collection of all $\alpha_n$, $\hat{r}_m$, $t_n$, and $\phi_{nm}$ for $n \in \{1, 2, ... \  N\}$ and $m \in \{1, 2, ... \  M\}$.
This is stored compactly in an $N \times M$ image, which we treat as the raw measurement produced by the sensor and which serves as the exteroceptive input to \ac{drtr}.

\subsection{Teach}\label{sec:teach}
\subsubsection{Odometry}
To build the topometric map, the system must estimate its local motion through odometry to capture the shape of the path.
For an incoming radar scan at time $t_{k}$, we wish to estimate $\mathbf{T}_{k-1,k} \in \text{SE}(2)$, the relative pose change that occurred since the previous frame at time $t_{k-1}$.
\Ac{drtr} achieves this using a re-implementation of the publicly available \acl{DRO} method \ac{DRO} \cite{legentil2025dro}.

Specifically, our implementation of \ac{DRO} estimates the body-centric velocity $\mathbf{v}_{k}$, assumed to be constant throughout a scan, and intra-frame orientation change $\mathbf{R}_k$ for each radar frame $k$.
The relative transform $\mathbf{T}_{k-1,k}$ is then computed by integrating the velocity forward in time.
All of our experiments use a gyroscope for an added source of interoceptive information.
When gyroscope information is available, \ac{DRO} directly preintegrates $\mathbf{R}_k$ from heading measurements.
We do not estimate the gyro bias as part of the state but update a bias using an exponential moving average whenever the robot's measured velocity is less than $0.05 \; \si{m/s}$.
Only a few stationary seconds are needed to provide a moderate odometry improvement from a biased gyroscope.

To estimate $\mathbf{v}_{k}$ (and $\mathbf{R}_k$ if no gyroscope is available), \ac{DRO} maximizes the cross-correlation between the entire `live' dense radar scan and a short history of dense scans.
The history of scans is maintained as a smoothed scan $\mathcal{M}$ that aggregates recent radar data in a motion- and Doppler-undistorted Cartesian image transformed into the current scan.
As the radar moves while recording each scan, the position of a given intensity $\psi_{nm}$ from the live scan changes with the pose of the radar at that time and the Doppler effect.
\Ac{DRO} computes to coordinates of $\psi_{nm}$ in $\mathcal{M}$, $(\Tilde{x}_{nm}, \Tilde{y}_{nm})$, correcting for Doppler and motion distortion as
\begin{equation}
    \left [ \begin{array}{c}
         \Tilde{x}_{nm}  \\
          \Tilde{y}_{nm}
    \end{array} \right ]
     = \mathbf{h}\left(\mathbf{v}_{k}, \mathbf{R}_k, \alpha_n, \hat{r}_m \right),
\end{equation}
with the details of the undistortion operator $\mathbf{h}(\cdot)$ left for the original \ac{DRO} paper for brevity \cite{legentil2025dro}.
The smoothed scan intensity is then interpolated at $(\Tilde{x}_{nm}, \Tilde{y}_{nm})$ using a bilinear interpolation function $\Gamma$ and compared with the incoming live scan.
The cross-correlation between the live and smoothed scans is then
\begin{equation}
    \Phi = \sum_{n=1}^N\sum_{m=1}^M \psi_{nm}\Gamma(\mathcal{M}, \Tilde{x}_{nm}, \Tilde{y}_{nm}).
\end{equation}
Gradient ascent is used to maximize $\Phi$\footnote{When the radar uses a specific modulation pattern, the objective function is supplemented with a specific Doppler constraint as in DRO \cite{legentil2025dro}.} with respect to $\mathbf{v}_{k}$ and, optionally, $\mathbf{R}_k$, recomputing $\Tilde{x}_{nm}$ and $\Tilde{y}_{nm}$ at each iteration.

After convergence, the smoothed scan is updated as an exponential moving average pixel-wise
\begin{equation}
    \mathcal{M}_{k} = (1-\beta)\mathcal{M}_{k-1} + \beta \mathcal{I}_{k},
\end{equation}
where $\beta$ is a tunable parameter between 0 and 1 and $\mathcal{I}_{k}$ is the post-optimization undistorted current intensity scan.
In \ac{drtr}, $\beta = 0.25$.

\subsubsection{Submaps}
\Ac{DRL} (Section~\ref{sec:repeat}) requires a dense intensity map of the environment to localize.
In Dr-BA \cite{lisus2026drba}, the method that introduced \ac{DRL}, the dense map is formed via a global bundle adjustment of \ac{DRO} smoothed scans.
As the motivation for \ac{drtr} is the wish to avoid averaging out perspective-dependent radar measurements, an effect inherent in global bundle adjustment, we directly use individual \textit{locally} smoothed scans as submaps.
Thus, if the algorithm decides to create a submap during frame $k$ of the teach pass, $\mathcal{M}_k$, a by-product of \ac{DRO}, is chosen.
Following Dr-BA, the stored representation of $\mathcal{M}_k$ is smoothed further and down-sampled into a constantly spaced voxel grid representation with a user-defined voxel size $\nu$, set to $\nu = 1.0\; \si{\meter}$ and $\nu = 0.5\;\si{\meter}$ for our on-road and off-road tests respectively.

The key question in topometric mapping is where to store a new submap.
For systems that have state-dependent biases, submap frequency is a bias-variance tradeoff \cite{tr_review_2026}.
Two common state-dependent artifacts of spinning radar are ground strike, which depends on unobservable pitch and roll, and multi-path returns, which often appear as weaker mirror images of structures.
Larger, less frequent submaps will average out random sensor noise in each scan, reducing variance.
Smaller, more frequent submaps will retain artifacts, increasing our bias.
In teach and repeat, where we attempt to localize from a state that matches the teach, the artifact bias from a live scan and the submap roughly cancel out, making the estimate of position relative to the teach more accurate.

In most previous versions of \ac{tr}, submaps are created at regular distance or heading intervals \cite{are_we_ready_for, rtr_1, hilger2026cfear}.
\ac{rtr} \cite{rtr_1} uses this approach with a spacing of $1.5 \si{\m}$ or $30^\circ$.
However, the use of dense radar scans enables a novel submap selection approach.
We propose using a similarity score between the incoming image and the most recent submap to decide whether to create a new submap.
This has been previously applied for radar-based loop-closure detection \cite{legentil2026drpogo, jang2023raplace}, but lends itself naturally to the idea that good submaps are those that contain unique viewpoint-dependent information and consistent radar artifacts.
The similarity score between two scans, $\mathcal{M}_i$ and $\mathcal{M}_j$, separated by an estimated pose $\mathbf{T}_{i,j}$, can be computed as a scaled cross-correlation $s$ \cite{legentil2026drpogo}
\begin{align}
    s &= \frac{g(\mathcal{M}_i, \mathcal{M}_j, \mathbf{T}_{i,j})}{\sum_{x,y} \mathcal{M}_j(x,y)},\label{eq:similarity_score}\\
    g(\mathcal{M}_i, \mathcal{M}_j, \mathbf{T}_{i,j}) &= \sum_{x,y} \Gamma\left(\mathcal{M}_i, \mathbf{T}_{i,j}\mathbf{p}_j\right) \mathcal{M}_j(x,y),\\
    \mathbf{p}_j &= \begin{bmatrix}x & y & 1\end{bmatrix}^\trans.
\end{align}
This approach provides a proxy for the overlap of the non-zero intensity measurements between the two scans.
A similarity score of 1.0 corresponds to an identical image, with less similar images decreasing in score down to 0.
The user thus only needs to tune a single intuitive parameter $\tau$, which creates a new submap if the similaity score between the previous submap scan and the current scan falls below $\tau$.
We use $s = 0.95$ for our offline and closed-loop results.
A study on the impact of using this similarity metric for submap selection is presented in Section~\ref{sec:submap_selection}.
The supplemental video shows an example of the similarity score changing as the robot drives and the scan changes, with Fig.~\ref{fig:garage_submapping} showing examples of automatically selected submaps from a \ac{drtr} teach.

\subsection{Repeat}\label{sec:repeat}

During a repeat, the system uses \acf{DRL} between the live smoothed scan and the nearest submap to estimate the robot's relative position to the path.
Consider a smoothed scan $\mathcal{M}_k$ at frame $k$ and a voxelized submap $\mathcal{S}$ with $P$ voxels within range of $\mathcal{M}_k$.
The objective function to localize pose $\mbf{T}_k$ in the submap is \cite{lisus2026drba}
\begin{align}
    \mbf{T}_k^\star &= \argmin_{\mbf{T}_k} \sum_{v = 1}^{P} w_{v} \left(\mathcal{S}_v - \Gamma(\mathcal{M}_k, \Tilde{x}_{v}, \Tilde{y}_{v})\right)^2,\label{eq:loc_objective}
\end{align}
where $\mathcal{S}_v$ is the intensity value at voxel $v$, $(\Tilde{x}_{v}, \Tilde{y}_{v})$ are the coordinates of voxel $v$ in $\mathcal{M}_k$, and $w_{v} > 0$ is the range and noise-dependent weight of each error.
\ac{DRO} still runs on every frame, providing an initial guess and prior for the \ac{DRL} optimization loop.

The bottom of Fig.~\ref{fig:garage_submapping} shows a snippet of a teach trajectory in black and repeat frames in blue that approximately repeat in the path of the teach.
The top of the figure shows example pairings of smoothed scans $\mathcal{M}_k$ and nearest submaps $\mathcal{S}$ that they are localized against.
Note the high overlap in view-dependent features as compared to the averaged Dr-BA map.

When closed-loop control is available, a \ac{MPC} guides the robot towards the taught path using the localization-derived relative pose.
The \ac{MPC} we use is identical to the one used in \cite{rtr_1}.

\section{Evaluation}
Two types of evaluation establish \ac{drtr} as a \ac{SOTA} localization approach using spinning radar: $344\;\si{\km}$ of on-road data and $20\;\si{\km}$ of off-road data.

For on-road evaluation, we use the Boreas-RT \cite{lisus_brrt26} dataset, which contains $643\;\si{\km}$ of data collected from a Navtech RAS6 spinning radar and a Silicon Sensing DMU41 \ac{IMU}.
The dataset covers multiple traversals of nine different routes, ranging from structured suburban environments to geometrically degenerate highways to unstructured rural farm roads.
Ground truth is provided using a post-processed Applanix POS-LV \ac{GNSS}-\ac{INS} system, and we use all available routes with ground truth errors below $3\;\si{\cm}$ to evaluate \ac{drtr}.
In total, around $58\;\si{\km}$ are used for mapping (teach) passes and $286\;\si{\km}$ are used to evaluate localization performance (repeat).
We use the same mapping/localization split as the baselines in the dataset paper \cite{lisus_brrt26}.

\begin{figure}[t]
    \centering
    \includegraphics[width=\linewidth]{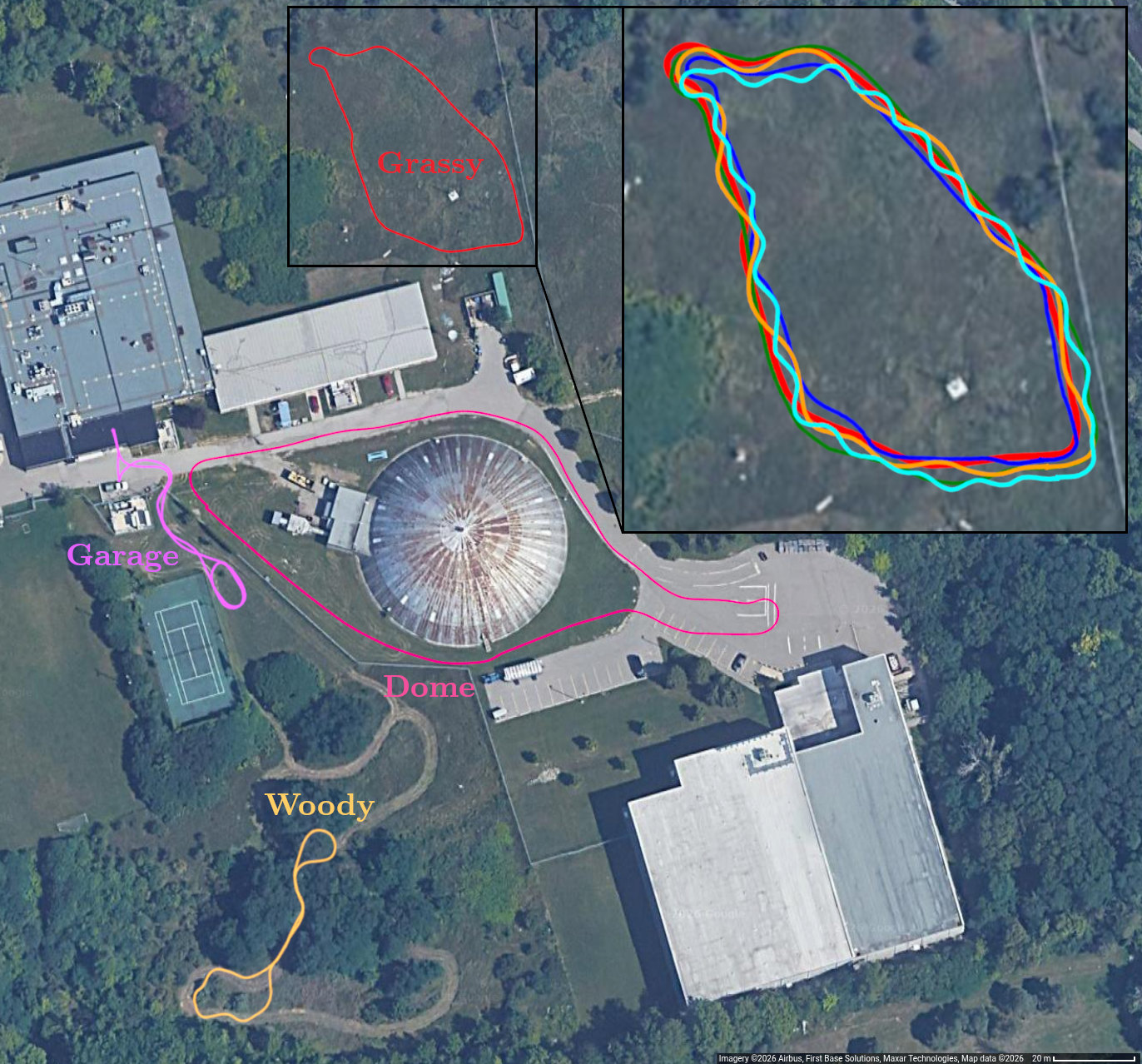}
    \caption{The four evaluation loops overlaid on a satellite image. Each loop contains five distinct manual recording sessions. The upper right inset illustrates all five for the \texttt{Grassy} sequence. Path-tracking errors are introduced to contrast with the autonomous repeats along the nominal path in the dataset.}
    \label{fig:dataset-loops}
\end{figure}

\begin{table}[t]
  \centering
  \caption{Self-collected Warthog off-road dataset summary. All sequences include radar, lidar, and IMU data.}
  \label{tab:dataset_summary}
  \begin{tabular}{l c c c c}
    \toprule
    \textbf{Sequence} & \textbf{Path Length} & \textbf{Manual} & \textbf{Autonomous} & \textbf{Total Dist.} \\
                      & \textbf{[m]}         & \textbf{Repeats}& \textbf{Repeats}    & \textbf{[km]} \\
    \midrule
    \texttt{Dome}   & 344 & 5 & 10 & 5.13 \\
    \texttt{Garage} & 168 & 5 & 10 & 2.52 \\
    \texttt{Grassy} & 168 & 5 & 10 & 2.52 \\
    \texttt{Woody}  & 143 & 5 & 10 & 2.15 \\
    \texttt{Campus}   & 1300 & 1 & 5 & 7.80 \\
    \bottomrule
  \end{tabular}
\end{table}

Additionally, we collected 20.1 km of new data with a Navtech RAS3 radar and InvenSense IAM-20680HT \ac{IMU} on the Clearpath Warthog in environments with less structure than in Boreas-RT.
Fig.~\ref{fig:dataset-loops} illustrates the four sequences used for evaluation, and Table~\ref{tab:dataset_summary} provides statistics on their distance and collection.
One third was collected manually and two-thirds autonomously.
The five manual runs are structured as follows: one nominal path, one to the left of nominal, one to the right of nominal, one low-frequency oscillation across the path, and one high-frequency oscillation across the path.
The inset of Fig.~\ref{fig:dataset-loops} shows all five manual runs for the \texttt{Grassy} sequence.
The \texttt{Campus} route does not include additional manual sequences as it was primarily meant to evaluate the long-term performance of \ac{drtr}.
The \texttt{Campus} path overlaps with the four smaller loops, so we omit it from Fig.~\ref{fig:dataset-loops} to prevent overlap.
Offline localization and autonomous repeats use the nominal path as the teach route.
Table~\ref{tab:dataset_summary} collates the total distances.
Off-road, the Warthog's \ac{GNSS} system suffers from significant position errors due to tree cover and other multi-path effects. 
As such, we record time-synchronized scans from an Ouster OS-1 128 beam lidar and use 2Fast-2Lamaa \cite{legentil_2026_2fast} localization to evaluate the vehicle's pose during each run.
This serves as our off-road ground truth.
We publicly release this dataset to facilitate future off-road radar localization developments by the community.

To evaluate localization, we compare \ac{drtr} with the existing \ac{SOTA} direct radar bundle adjustment (Dr-BA) \cite{lisus2026drba}, which uses direct methods on a globally consistent map.
Additionally, we quantify the performance of the previous \ac{ICP} \ac{rtr} to provide context for the closed-loop evaluation.
In closed loop, we compare to \ac{rtr} \cite{rtr_1} as the previous closed-loop \ac{SOTA} method and also run \ac{ltr} \cite{are_we_ready_for} to compare to a \ac{SOTA} closed-loop lidar method.

\begin{table}[t]
    \centering
    \caption{Offline RMS localization errors for RT\&R, Dr-BA, DRT\&R, and 2Fast-2Lamaa on the Boreas-RT dataset. \\Best radar-based results are bolded.}
    \setlength{\tabcolsep}{4pt}
    \begin{tabularx}{\linewidth}{lTTTT}
        \toprule
        & \textbf{RT\&R} \scriptsize\cite{rtr_1}
        & \textbf{Dr-BA} \scriptsize\cite{lisus2026drba}
        & \textbf{DRT\&R} \scriptsize(ours)
        & \textcolor{gray}{\textbf{2Fast \cite{legentil_2026_2fast}}} \\
        \midrule
        \multicolumn{5}{l}{\texttt{Suburbs}} \\
        \quad Long. [\si{\m}]
        & 0.097 & 0.047 & \textbf{0.032} & \textcolor{gray}{0.028}\\
        \quad Lat. [\si{\m}]
        & 0.065 & 0.059 & \textbf{0.039} & \textcolor{gray}{0.026}\\
        \quad Yaw [$^\circ$]
        & 0.124 & 0.067 & \textbf{0.055} & \textcolor{gray}{0.029}\\
        \midrule
        \multicolumn{5}{l}{\texttt{Industrial}} \\
        \quad Long. [\si{\m}]
        & 0.081 & 0.045 & \textbf{0.030} & \textcolor{gray}{0.023}\\
        \quad Lat. [\si{\m}]
        & 0.054 & 0.055 & \textbf{0.038} & \textcolor{gray}{0.022}\\
        \quad Yaw [$^\circ$]
        & 0.117 & 0.055 & \textbf{0.053} & \textcolor{gray}{0.034}\\
        \midrule
        \multicolumn{5}{l}{\texttt{Tunnel}} \\
        \quad Long. [\si{\m}]
        & 2.030$^{\frac{4}{8}}$ & 0.131 & \textbf{0.056} & \textcolor{gray}{0.127}\\
        \quad Lat. [\si{\m}]
        & 0.144$^{\frac{4}{8}}$ & 0.125 & \textbf{0.058} & \textcolor{gray}{0.032}\\
        \quad Yaw [$^\circ$]
        & 0.172$^{\frac{4}{8}}$ & 0.128 & \textbf{0.064} & \textcolor{gray}{0.031}\\
        \midrule
        \multicolumn{5}{l}{\texttt{Skyway}} \\
        \quad Long. [\si{\m}]
        & -$^{\frac{4}{4}}$ & 0.291 & \textbf{0.075} & \textcolor{gray}{0.082}\\
        \quad Lat. [\si{\m}]
        & -$^{\frac{4}{4}}$ & 0.222 & \textbf{0.071} & \textcolor{gray}{0.038}\\
        \quad Yaw [$^\circ$]
        & -$^{\frac{4}{4}}$ & 0.151 & \textbf{0.065} & \textcolor{gray}{0.048}\\
        \midrule
        \multicolumn{5}{l}{\texttt{Regional}} \\
        \quad Long. [\si{\m}]
        & 0.273 & 0.065 & \textbf{0.041} & \textcolor{gray}{0.033}\\
        \quad Lat. [\si{\m}]
        & 0.072 & 0.082 & \textbf{0.067} & \textcolor{gray}{0.032}\\
        \quad Yaw [$^\circ$]
        & 0.121 & 0.064 & \textbf{0.062} & \textcolor{gray}{0.029}\\
        \midrule
        \multicolumn{5}{l}{\texttt{Farm}} \\
        \quad Long. [\si{\m}]
        & 0.204$^{\frac{6}{9}}$ & 0.302$^{\frac{1}{9}}$ & \textbf{0.076} & \textcolor{gray}{0.065}\\
        \quad Lat. [\si{\m}]
        & 0.212$^{\frac{6}{9}}$ & 0.481$^{\frac{1}{9}}$ & \textbf{0.120} & \textcolor{gray}{0.055}\\
        \quad Yaw [$^\circ$]
        & 0.166$^{\frac{6}{9}}$ & 0.274$^{\frac{1}{9}}$ & \textbf{0.132} & \textcolor{gray}{0.045}\\
        \bottomrule
        \multicolumn{5}{l}{\scriptsize Superscript indicates number failed over the total number of sequences.}
    \end{tabularx}
    \label{tab:localization_boreas}
\end{table}
\section{Results}
\begin{figure}[t!]
    \centering
    \resizebox{\linewidth}{!}{\input{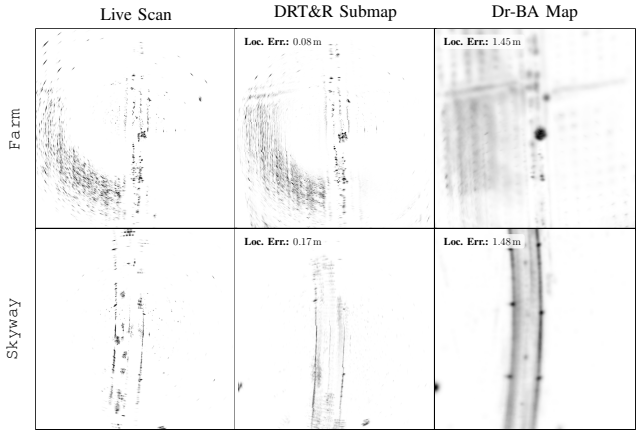}}
    \caption{ Example regions of \texttt{Farm} (top) and \texttt{Skyway} (bottom) sequences where localizing to local submaps yields a sizeable improvement in translational error. Left to right: the live scan being used to localize, the DRT\&R submap being localized against in our method, a section of the global map being used to localize for Dr-BA. \texttt{Farm} localization improvement can be attributed to a locally re-observed ground strike and a perspective-dependent view of the tree in the middle of the scan. \texttt{Skyway} localization improvement can be attributed to a better overlap in the visibility of the road.}
    \label{fig:challenging_examples}
\end{figure}

Overall, \ac{drtr} achieves \ac{SOTA} radar localization performance on all on-road and off-road data. 
Table~\ref{tab:localization_boreas} aggregates the SE(2) pose errors for all considered Boreas-RT routes.
The fourth column (2Fast) contextualizes the localization quality by comparing to the current \ac{SOTA} 3D lidar method, 2Fast-2Lamaa \cite{legentil_2026_2fast}, on the same sequences. 
The largest improvement over previous radar methods occurs in the \texttt{Farm} sequence, where position errors drop by 74.9\%, followed by improvements on the geometrically degenerate \texttt{Skyway} and \texttt{Tunnel} routes.
Fig.~\ref{fig:challenging_examples} illustrates an example where a consistent ground strike in the \texttt{Farm} route due to vehicle roll provides a strong localization feature but is removed from Dr-BA's global map, since the region is clear from other viewpoints.
Fig.~\ref{fig:challenging_examples} also shows a region of the \texttt{Skyway} where the local curvature prevents the vehicle from seeing too far ahead, which in turn prevents the localization estimate from sliding along the walls as it does in the global map.
In the \texttt{Tunnel}, longitudinal position errors are reduced, even beyond those of the lidar, likely on account of the available Doppler information leveraged by \ac{DRO}.
The \texttt{Tunnel} radar scans contain multiple reflections between the cement walls of the tunnel, which, although non-physical, provide additional features at longer observed range and improve the quality of the orientation estimate by 50\% over other radar methods.

Localization off-road follows the same trend as on-road: \ac{drtr} decreases position error on average by 50\% and orientation error by 32.2\% compared to Dr-BA.
In Table~\ref{tab:localization_warthog}, \ac{drtr} has lower localization errors on all sequences.
The largest improvement of 68.5\% occurs on the \texttt{Campus} sequence.
We observe that regions with multiple observations from topologically distant but physically close locations along the path lead to imprecise Dr-BA global maps.
Localization in these areas is much worse for Dr-BA than our topometric approach, in which the submaps for topologically distant places share no information.

\subsection{Effect of Submap Frequency on Localization}
\label{sec:submap_selection}

\begin{figure}
    \centering
    \includegraphics[width=0.98\linewidth]{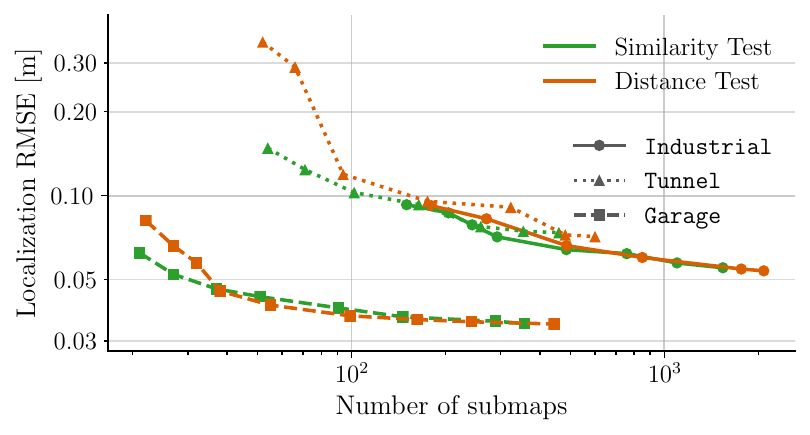}
    \caption{A study of the impact of the number of submaps on localization accuracy. Two submap creation methods are considered: standard distance-based and our new similarity-based approach. Results are presented on the \texttt{Industrial}, \texttt{Tunnel}, and \texttt{Garage} sequences, which represent well-structured, geometrically degenerate, and off-road environments, respectively. Performance is comparable between the two approaches until thresholds are set to yield few submaps. When submaps are created infrequently, the similarity-based approach is superior.}
    \label{fig:submap_tradeoff}
\end{figure}

\begin{figure}
    \centering
    \includegraphics[width=0.95\linewidth]{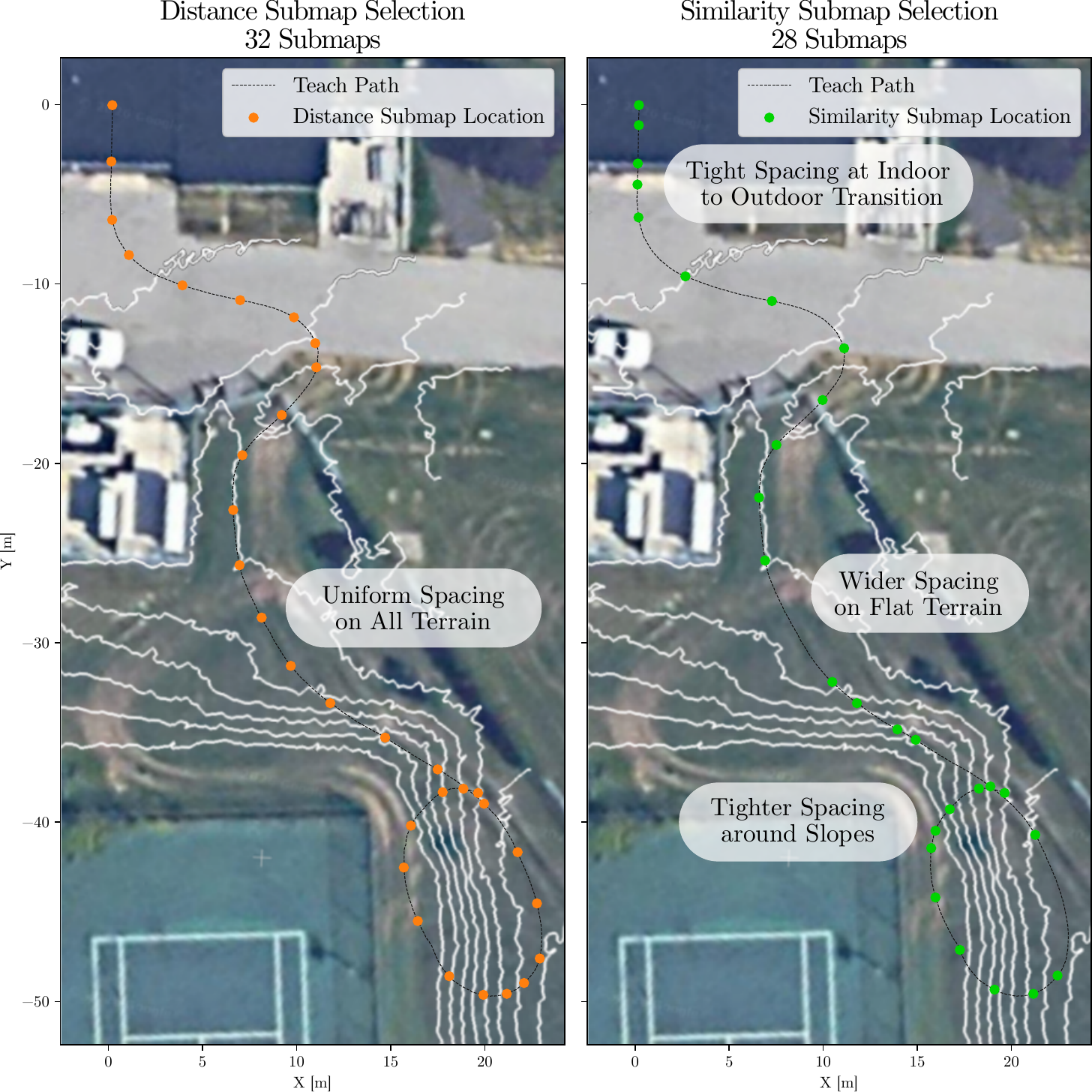}
    \caption{A comparison of submap locations using a distance of $3.0\; \si{\m}$ or a similarity threshold of 0.8. Parameters generate fewer submaps than in the actual experiments for visual clarity. Contour elevation lines estimated by the lidar, overlaid in white, illustrate the slopes along the route. On this subset of the \texttt{Garage} sequence, places with rapid changes in the pitch or roll of the vehicle have a higher density of submaps than flat regions.} 
    \label{fig:submap_comparison}
    \vspace{-3mm}
\end{figure}

To study the impact of the similarity threshold while controlling for the total number of submaps it produces, we compare the two submap tests as a function of the number of submaps they create.
We observe a similar effect for both tests, visualized in Fig.~\ref{fig:submap_tradeoff}: more submaps lead to more accurate localization.
As localization improves, the storage size of the topometric map grows with the number of submaps.
In the \texttt{Garage}, the performance gains between 40 submaps and 400 are very small, but 10 times the disk space is required.
When fewer submaps are desirable, similarity thresholding becomes valuable.
This is most pronounced in the \texttt{Tunnel}.
At the outside-to-inside transition at the beginning of the tunnel, similarity scoring detects a sudden scene change as the radar becomes occluded in the entrance, triggering three closely spaced submaps during the transition.
Distance thresholding does not capture the transition, and localization suffers as the vehicle enters into the tunnel.
In the \texttt{Garage},  there is a similar transition as the robot drives outside.
In the top of Fig.~\ref{fig:submap_comparison}, two submaps wrap around this transition.
Fig.~\ref{fig:submap_comparison} also demonstrates the value of similarity submapping as the robot begins to pitch and roll.
As the robot starts to climb a hill, frequent submaps capture the changing radar ground strike in this environment.

\subsection{Effect of Path Following on Localization}
\begin{figure}
    \centering
    \includegraphics[width=0.85\linewidth]{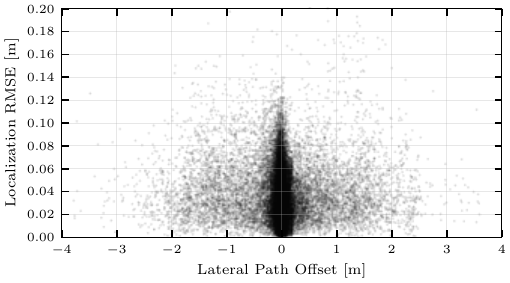}
    \caption{A per-frame comparison of the localization position error to the lateral offset from the teach path. We do not see a correlation, which validates that \ac{DRL} is robust even if path-tracking performance degrades.}
    \label{fig:loc_e_vs_pte}
    \vspace{-3mm}
\end{figure}

For the Warthog off-road data, most manually piloted runs are purposely driven to the left or right of the path, or with high-frequency oscillations characteristic of a poorly tuned controller.
Fig.~\ref{fig:loc_e_vs_pte} plots the relationship between the norm of the localization position error and the distance from the taught path.
Within our dataset range of $4\;\si{\m}$ of lateral offset from the path, there is no correlation between the distance from the path and the results obtained by \ac{DRL}.
This suggests that the localization remains robust to temporary deviations from the path or poor initialization.

\begin{table}[t]
    \centering
    \caption{Offline RMS localization errors for \ac{rtr}, Dr-BA, and \ac{drtr} on the self-collected Warthog dataset.}
    \setlength{\tabcolsep}{4pt}
    \begin{tabularx}{\linewidth}{llOOO}
        \toprule
        & & \textbf{RT\&R} \scriptsize\cite{rtr_1}
        & \textbf{Dr-BA} \scriptsize\cite{lisus2026drba}
        & \textbf{DRT\&R} \scriptsize(ours) \\
        \midrule
        \multirow{3}{*}{\texttt{Dome}}
        & Long. [\si{\m}]
        & 0.058 & 0.041 & \textbf{0.029} \\
        & Lat. [\si{\m}]
        & 0.040 & 0.046 & \textbf{0.025} \\
        & Yaw [$^\circ$]
        & 0.354 & 0.182 & \textbf{0.155} \\
        \midrule
        \multirow{3}{*}{\texttt{Garage}}
        & Long. [\si{\m}]
        & 0.068 & 0.045 & \textbf{0.023} \\
        & Lat. [\si{\m}]
        & 0.054 & 0.040 & \textbf{0.021} \\
        & Yaw [$^\circ$]
        & 0.433 & 0.303 & \textbf{0.193} \\
        \midrule
        \multirow{3}{*}{\texttt{Grassy}}
        & Long. [\si{\m}]
        & 0.130 & 0.066 & \textbf{0.028} \\
        & Lat. [\si{\m}]
        & 0.087 & 0.054 & \textbf{0.031} \\
        & Yaw [$^\circ$]
        & 0.517 & 0.300 & \textbf{0.167} \\
        \midrule
        \multirow{3}{*}{\texttt{Woody}}
        & Long. [\si{\m}]
        & 0.209 & 0.050 & \textbf{0.028} \\
        & Lat. [\si{\m}]
        & 0.124 & 0.056 & \textbf{0.027} \\
        & Yaw [$^\circ$]
        & 0.708 & 0.290 & \textbf{0.214} \\
        \midrule
        \multirow{3}{*}{\texttt{Campus}}
        & Long. [\si{\m}]
        & 0.112 & 0.108 & \textbf{0.034} \\
        & Lat. [\si{\m}]
        & 0.059 & 0.092 & \textbf{0.029} \\
        & Yaw [$^\circ$]
        & 0.394 & 0.287 & \textbf{0.173}\\
        \bottomrule
    \end{tabularx}
    \label{tab:localization_warthog}
\end{table}

\begin{table}[t]
    \centering
    \caption{Online path-tracking errors for RT\&R, DRT\&R, and LT\&R on the self-collected Warthog dataset.}
    \setlength{\tabcolsep}{4pt}
    \begin{tabularx}{\linewidth}{llOOO}
        \toprule
        & & \textbf{RT\&R} \scriptsize\cite{rtr_1}
        & \textbf{DRT\&R} \scriptsize(ours) 
        & \textcolor{gray}{\textbf{LT\&R} \cite{are_we_ready_for}} \\
        \midrule
        \multirow{2}{*}{\texttt{Dome}}
        & RMS [\si{\m}]
        & 0.050 & \textbf{0.038} & \textcolor{gray}{0.046}\\
        & Max. [\si{\m}]
        & 0.212 & \textbf{0.168} & \textcolor{gray}{0.184} \\
        \midrule
        \multirow{2}{*}{\texttt{Garage}}
        & RMS [\si{\m}]
        & 0.054 & \textbf{0.051} & \textcolor{gray}{0.048}\\
        & Max. [\si{\m}]
        & 0.300 & \textbf{0.184} & \textcolor{gray}{0.166}\\
        \midrule
        \multirow{2}{*}{\texttt{Grassy}}
        & RMS [\si{\m}]
        & 0.118 & \textbf{0.060} & \textcolor{gray}{0.067}\\
        & Max. [\si{\m}]
        & 0.549 & \textbf{0.349} & \textcolor{gray}{0.227} \\
        \midrule
        \multirow{2}{*}{\texttt{Woody}}
        & RMS [\si{\m}]
        & 0.130 & \textbf{0.048} & \textcolor{gray}{0.040}\\
        & Max. [\si{\m}]
        & 0.546 & \textbf{0.199} & \textcolor{gray}{0.143}\\
        \midrule
        \multirow{2}{*}{\texttt{Campus}}
        & RMS [\si{\m}]
        & --- & \textbf{0.062} & \textcolor{gray}{---}\\
        & Max. [\si{\m}]
        & --- & \textbf{0.403} & \textcolor{gray}{---}\\
        \bottomrule
    \end{tabularx}
        \vspace{-3mm}

    \label{tab:path_tracking_warthog}
\end{table}

\subsection{Closed-Loop Path Tracking}
The key validation for \ac{drtr} is the integration of the full odometry, mapping, and localization stack onto the Clearpath Warthog system. 
To compare \ac{rtr} and \ac{drtr}, the nominal manual sequence from the Warthog dataset is used as the teach.
The Warthog repeats at $1\;\si{\meter/\second}$ along all five routes described in Table~\ref{tab:dataset_summary} five times using both \ac{rtr} and \ac{drtr} for its state estimation for a total evaluation distance of $10.6\; \si{\km}$.
For additional context, one \ac{ltr} repeat of each route shows the controller's path-tracking performance with more accurate localization.
An identical \ac{MPC} selects velocity commands to minimize lateral path-tracking error for all three cases.

Table~\ref{tab:path_tracking_warthog} demonstrates a 44\% average reduction in path-tracking error using \ac{drtr} compared to \ac{rtr}.
The Warthog routes in Tables \ref{tab:localization_warthog} and \ref{tab:path_tracking_warthog} are sorted from most to least structured.
For \ac{rtr}, this correlates with increases in localization and path-tracking error; however, for \ac{drtr}, the localization \ac{RMSE} remains consistently around $3.5\; \si{\cm}$ and path-tracking \ac{RMSE} ranges between $3.8\;\si{\cm}$ and $6.2\; \si{\cm}$.
More importantly, the maximum path-tracking errors decrease in all sequences, with the largest reduction of $35\; \si{\cm}$ in the \texttt{Woody} sequence.
We run \ac{drtr} for an additional $6.5\;\si{\km}$ of autonomous repeats across the \texttt{Campus} route, which includes the four smaller sequences and more of the parking lot in a continuous loop.
Over this $6.5\; \si{\km}$ of testing, the \ac{RMS} path-tracking error was $6.2\; \si{\cm}$.

Comparing \ac{drtr} to \ac{ltr}, path-tracking errors are within $1\;\si{\cm}$ of each other, which is insignificant relative to the size of the vehicle.
In our tests, spinning radar performance matches lidar for autonomous navigation tasks.
Over $10.3\;\si{\km}$ of autonomous repeating, \ac{drtr} exhibited smooth, reliable path tracking, naturally handling three-dimensional terrain and abrupt indoor-outdoor transitions.

Both \ac{rtr} and \ac{drtr} run in real time (faster than the $4\; \si{Hz}$ radar frame rate) on a laptop with an Intel i7-12800H and NVIDIA RTX A4500 GPU.
In the \texttt{Woody} sequence, \ac{rtr} takes $38.98\; \si{ms}$ for odometry and $5.80\; \si{ms}$ for localization;
\ac{drtr} takes $107.67\; \si{ms}$ for odometry and $55.58\; \si{ms}$ for localization.
The \ac{DRO} implementation is GPU-accelerated.
In this sequence, \ac{rtr} runs especially quickly because the point-cloud extraction finds few points. 
However, this low point density also contributes to the low-quality localization performance in this location.
We leave code optimization of \ac{DRO} for future work.

\section{Conclusion}
We present \acf{drtr}: the first closed-loop navigation system using direct methods for spinning radar.
\ac{drtr} demonstrates state-of-the-art localization performance on $344\;\si{\km}$ of data from the on-road Boreas-RT dataset and on $20\;\si{\km}$ of data from our self-collected off-road dataset.
It also demonstrates $6\;\si{\cm}$ \ac{RMS} path-tracking errors driving a Clearpath Warthog off-road for more than $10\;\si{\km}$ in closed loop, reducing the average path-tracking error by 44\% by using direct methods rather than point clouds.
We validate a core benefit of topometric approaches to localization: storing small local maps preserves state-dependent artifacts such as multi-path reflections and ground strikes that improve localization.
Additionally, we propose an improved method for deciding when to create a new submap based on the similarity of an incoming scan and the active submap.
This approach yields intuitive benefits, increasing density around transitional locations and saving on storage space in consistent environments.
We look forward to exploring the limits of this approach in more challenging field tests in the future.


{\footnotesize
\bibliographystyle{IEEEtranN}
\bibliography{references}}

@string{tro      = {IEEE Trans. on Robotics} }

@string{arxiv   = {arXiv preprint} }

@string{corl    = {Conference on Robot Learning (CoRL)} }

@string{crv     = {Proc.~of the Conf.~on Robots and Vision} }

@string{icra    = {Proc.~of the IEEE Intl.~Conf.~on Robotics \& Automation 
(ICRA)} }

@string{ijrr    = {Intl.~Journal~of Robotics Research (IJRR)} }

@string{iros    = {Proc.~of the IEEE/RSJ Intl.~Conf.~on Intelligent Robots and 
Systems (IROS)} }

@string{iv      = {Proc.~of the IEEE Intelligent Vehicles Symposium (IV)} }

@string{ral     = {IEEE Robotics and Automation Letters (RA-L)} }

@string{rss     = {Proc.~of Robotics: Science and Systems (RSS)} }

@string{springer= {Springer Verlag} }

@string{tro     = {IEEE Trans.~on Robotics (TRO)} }

@string{tiv     = {IEEE Trans.~on Intelligent Vehicles (T-IV)} }

@article{are_we_ready_for,
  title={{Are We Ready for Radar to Replace Lidar in All-weather Mapping and Localization?}},
  author={Burnett, Keenan and Wu, Yuchen and Yoon, David J and Schoellig, Angela P and Barfoot, Timothy D},
  journal=ral,
  volume={7},
  number={4},
  pages={10328--10335},
  year={2022},
  publisher={IEEE}
}

@article{finer_points,
	author = {Preston-Krebs, Elliot and Lisus, Daniil and Barfoot, Timothy D.},
	journal = crv,
	year = {2025},
	month = may,
	publisher = {},
	title = {The {Finer} {Points}: A {Systematic} {Comparison} of {Point}-{Cloud} {Extractors} for {Radar} {Odometry}},
}

@article{a_new_wave_radar,
  title={{A New Wave in Robotics: Survey on Recent mmWave Radar Applications in Robotics}},
  author={Harlow, Kyle and Jang, Hyesu and Barfoot, Timothy D and Kim, Ayoung and Heckman, Christoffer},
  journal={IEEE Trans. Robot.},
  year={2024},
  publisher={IEEE}
}

@article{venon2022millimeter,
  title={{Millimeter Wave FMCW Radars for Perception, Recognition and Localization in Automotive Applications: A Survey}},
  author={Venon, Arthur and Dupuis, Yohan and Vasseur, Pascal and Merriaux, Pierre},
  journal={IEEE Trans. Intell. Vehicles},
  volume={7},
  number={3},
  pages={533--555},
  year={2022},
  publisher={IEEE}
}

@inproceedings{under_the_radar,
  author = {Dan Barnes and Ingmar Posner},
  title = {Under the Radar: Learning to Predict Robust Keypoints for Odometry Estimation and Metric Localisation in Radar},
  booktitle=icra,
  year = {2020}
}

@INPROCEEDINGS{adolfsson2021CFEAR,
author = {Adolfsson, Daniel and Magnusson, Martin and Alhashimi, Anas and Lilienthal, Achim and Andreasson, Henrik},
year = {2021},
month = {09},
pages = {5462-5469},
title = {{CFEAR Radarodometry - Conservative Filtering for Efficient and Accurate Radar Odometry}},
booktitle=iros,
doi = {10.1109/IROS51168.2021.9636253}
}

@INPROCEEDINGS{cen2019,
  author={Cen, Sarah H. and Newman, Paul},
  booktitle=icra, 
  title={{Radar-only Ego-motion Estimation in Difficult Settings via Graph Matching}}, 
  year={2019},
  volume={},
  number={},
  pages={298-304},
  doi={10.1109/ICRA.2019.8793990}
}

@article{hong2022radarslam,
author = {Ziyang Hong and Yvan Petillot and Andrew Wallace and Sen Wang},
title ={{RadarSLAM: A Robust Simultaneous Localization and Mapping System for All Weather Conditions}},
journal = ijrr,
volume = {41},
number = {5},
pages = {519-542},
year = {2022},
doi = {10.1177/02783649221080483},
eprint = {
        https://doi.org/10.1177/02783649221080483
}
}

@ARTICLE{callmer2011radarslamusingvisualfeatures,
author = {Callmer, Jonas and Törnqvist, David and Gustafsson, Fredrik and Svensson, Henrik and Carlbom, Pelle},
year = {2011},
month = {12},
pages = {},
title = {{Radar SLAM Using Visual Features}},
volume = {2011},
journal = {EURASIP Journal on Advances in Signal Processing},
doi = {10.1186/1687-6180-2011-71}
}

@article{VTR,
  title={{Visual Teach and Repeat for Long-Range Rover Autonomy}},
  author={Furgale, Paul and Barfoot, Timothy D},
  journal={J. Field Robot.},
  volume={27},
  number={5},
  pages={534--560},
  year={2010},
  publisher={Wiley Online Library}
}

@inproceedings{lisus2025pointing,
  author={Lisus, Daniil and Laconte, Johann and Burnett, Keenan and Zhang, Ziyu and Barfoot, Timothy D.},
  title={{Pointing the Way: Refining Radar-Lidar Localization Using Learned ICP Weights}},
  booktitle=crv,
  year={2025}
}

@inproceedings{masking_by_moving,
  author = {Barnes, Dan and Weston, Rob and Posner, Ingmar},
  title = {{Masking by Moving: Learning Distraction-Free Radar Odometry from Pose Information}},
  booktitle = {Conference on Robot Learning (CoRL)},
  year = {2019}
}

@inproceedings{lisus2026drba,
  author={Lisus, Daniil and Le Gentil, Cedric and Barfoot, Timothy D.},
  title={{Dr-BA: Separable Optimization for Direct Radar Bundle Adjustment \& Localization}},
  booktitle=rss,
  year={2026}
}

@inproceedings{legentil2025dro,
  title={{DRO: Doppler-Aware Direct Radar Odometry}},
  author={Le Gentil, Cedric and Brizi, Leonardo and Lisus, Daniil and Qiao, Xinyuan and Grisetti, Giorgio and Barfoot, Timothy D.},
  booktitle=rss,
  year={2025}
}

@inproceedings{legentil2026drpogo,
  title={{Dr-PoGO: Direct Radar Pose-Graph Optimization}},
  author={Le Gentil, Cedric and Li, Weican and Brizi, Leonardo and Barfoot, Timothy D.},
  booktitle=icra,
  year={2026}
}

@ARTICLE{adolfsson2023tbv,
  author={Adolfsson, Daniel and Karlsson, Mattias and Kubelka, Vladimír and Magnusson, Martin and Andreasson, Henrik},
  journal=ral,
  title={{TBV Radar SLAM -- Trust but Verify Loop Candidates}},
  year={2023},
  volume={8},
  number={6},
  pages={3613-3620},
  doi={10.1109/LRA.2023.3268040}
}

@article{lisus_brrt26,
  title = {{Boreas Road Trip: {{A}} Multi-Sensor Autonomous Driving Dataset on Challenging Roads}},
  shorttitle = {Boreas Road Trip},
  author = {Lisus, Daniil and Papais, Katya M. and Le Gentil, Cedric and {Preston-Krebs}, Elliot and Lambert, Andrew and Leung, Keith Y. K. and Barfoot, Timothy D.},
  year = 2026,
  month = sep,
  journal = {The International Journal of Robotics Research},
  pages = {02783649261479362},
  publisher = {SAGE Publications Ltd STM},
  issn = {0278-3649},
  doi = {10.1177/02783649261479362},
  urldate = {2026-09-12},
  langid = {english}
}

@article{legentil_2026_2fast,
  title = {{{{2Fast-2Lamaa}}: {{Large-scale}} Lidar-Inertial Localization and Mapping with Continuous Distance Fields}},
  shorttitle = {{{2Fast-2Lamaa}}},
  author = {Le Gentil, Cedric and Falque, Raphael and Lisus, Daniil and Barfoot, Timothy D.},
  year = 2026,
  month = jun,
  journal = {The International Journal of Robotics Research},
  pages = {02783649261451003},
  publisher = {SAGE Publications Ltd STM},
  issn = {0278-3649},
  doi = {10.1177/02783649261451003},
  urldate = {2026-09-12},
  langid = {english}
}

@ARTICLE{nader2024survey,
  author={Abu-Alrub, Nader J. and Rawashdeh, Nathir A.},
  journal=tiv, 
  title={{Radar Odometry for Autonomous Ground Vehicles: A Survey of Methods and Datasets}}, 
  year={2024},
  volume={9},
  number={3},
  pages={4275-4291},
  doi={10.1109/TIV.2023.3340513}
}

@ARTICLE{adolfsson2023cfear,
  author={Adolfsson, Daniel and Magnusson, Martin and Alhashimi, Anas and Lilienthal, Achim J. and Andreasson, Henrik},
  journal=tro, 
  title={{Lidar-Level Localization With Radar? The CFEAR Approach to Accurate, Fast, and Robust Large-Scale Radar Odometry in Diverse Environments}}, 
  year={2023},
  volume={39},
  number={2},
  pages={1476-1495},
  doi={10.1109/TRO.2022.3221302}
}

@INPROCEEDINGS{park2020pharao,
  author={Park, Yeong Sang and Shin, Young-Sik and Kim, Ayoung},
  booktitle=icra, 
  title={{PhaRaO: Direct Radar Odometry using Phase Correlation}}, 
  year={2020},
  volume={},
  number={},
  pages={2617-2623},
  doi={10.1109/ICRA40945.2020.9197231}
}

@INPROCEEDINGS{holder2019realtime,
  author={Holder, Martin and Hellwig, Sven and Winner, Hermann},
  booktitle={2019 IEEE Intelligent Vehicles Symposium (IV)}, 
  title={{Real-Time Pose Graph SLAM based on Radar}}, 
  year={2019},
  volume={},
  number={},
  pages={1145-1151},
  doi={10.1109/IVS.2019.8813841}}

@ARTICLE{kim2022scancontextpp,
  author={Kim, Giseop and Choi, Sunwook and Kim, Ayoung},
  journal=tro, 
  title={{Scan Context++: Structural Place Recognition Robust to Rotation and Lateral Variations in Urban Environments}}, 
  year={2022},
  volume={38},
  number={3},
  pages={1856-1874},
  doi={10.1109/TRO.2021.3116424}}

@INPROCEEDINGS{jang2023raplace,
  author={Jang, Hyesu and Jung, Minwoo and Kim, Ayoung},
  booktitle=iros, 
  title={{RaPlace: Place Recognition for Imaging Radar using Radon Transform and Mutable Threshold}}, 
  year={2023},
  volume={},
  number={},
  pages={11194-11201},
  doi={10.1109/IROS55552.2023.10341883}
}

@article{jang2025xpress,
  title={{XPRESS: X-Band Radar Place Recognition via Elliptical Scan Shaping}},
  author={Jang, Hyesu and Yang, Wooseong and Kim, Ayoung and Lee, Dongje and Kim, Hanguen},
  journal={IEEE Robotics and Automation Letters},
  volume={10},
  number={12},
  pages={13121--13128},
  year={2025},
  publisher={IEEE}
}

@INPROCEEDINGS{rtr_1,
  author={Qiao, Xinyuan and Krawciw, Alexander and Lilge, Sven and Barfoot, Timothy D.},
  booktitle={2025 IEEE International Conference on Robotics and Automation (ICRA)}, 
  title={Radar Teach and Repeat: Architecture and Initial Field Testing}, 
  year={2025},
  volume={},
  number={},
  pages={13021-13027},
  doi={10.1109/ICRA55743.2025.11128412}
}

@article{tr_review_2026,
   author = "Krawciw, Alexander and Barfoot, Timothy D.",
   title = "{Local Maps Are All You Need: A Review of Topometric Teach and Repeat Navigation}", 
   journal= "Annual Review of Control, Robotics, and Autonomous Systems",
   year = "2026",
   volume = "9",
   number = "Volume 9, 2026",
   pages = "301-324",
   doi = "https://doi.org/10.1146/annurev-control-032724-020548",
   publisher = "Annual Reviews",
   issn = "2573-5144",
}

@ARTICLE{tro_xiao_2026,
  author={Xiao, Renxiang and Chen, Yichen and Zhang, Yuanfan and Shao, Qianyi and Chen, Yushuai and Han, Yuxuan and Lou, Yunjiang and Hu, Liang},
  journal={IEEE Transactions on Robotics}, 
  title={{LiDAR Teach, Radar Repeat: Robust Cross-Modal Navigation in Degenerate and Varying Environments}}, 
  year={2026},
  volume={42},
  number={},
  pages={2500-2520},
  doi={10.1109/TRO.2026.3699433}
}

@misc{boxan_2026,
  title = {{One Year in a Forest: {{Analyzing}} the Challenges of Autonomous Navigation in Subarctic Environments}},
  shorttitle = {One Year in a Forest},
  author = {Boxan, Mat{\v e}j and Lauzon, Nicolas and Vannini, Veronica and {Turgeon-Roy}, Mathis and Pomerleau, Fran{\c c}ois},
  year = 2026,
  month = aug,
  note = {arXiv:2608.27628},
  eprint = {2608.27628},
  primaryclass = {cs.RO},
  publisher = {arXiv},
  doi = {10.48550/arXiv.2608.27628},
  urldate = {2026-09-09},
  archiveprefix = {arXiv}
}

@misc{kolhe_2026,
  title = {Pushing {{Radar Odometry Beyond}} the {{Pavement}}: {{Current Capabilities}} and {{Challenges}}},
  shorttitle = {Pushing {{Radar Odometry Beyond}} the {{Pavement}}},
  author = {Kolhe, Shaunak and Jiang, Peng and Wigness, Maggie and Osteen, Philip and Overbye, Timothy and Ellis, Chrisitan and Saripalli, Srikanth},
  year = 2026,
  month = apr,
  note = {arXiv:2604.24674},
  eprint = {2604.24674},
  primaryclass = {cs.RO},
  publisher = {arXiv},
  doi = {10.48550/arXiv.2604.24674},
  urldate = {2026-09-09},
  archiveprefix = {arXiv}
}

@misc{fomo_dataset,
  title = {{{FoMo}}: {{A Multi-Season Dataset}} for {{Robot Navigation}} in {{For\^et Montmorency}}},
  shorttitle = {{{FoMo}}},
  author = {Boxan, Mat{\v e}j and Jeanson, Gabriel and Krawciw, Alexander and Daum, Effie and Qiao, Xinyuan and Lilge, Sven and Barfoot, Timothy D. and Pomerleau, Fran{\c c}ois},
  year = 2026,
  month = mar,
  note = {arXiv:2603.08433},
  eprint = {2603.08433},
  primaryclass = {cs.RO},
  publisher = {arXiv},
  doi = {10.48550/arXiv.2603.08433},
  urldate = {2026-09-09},
  archiveprefix = {arXiv}
}

@inproceedings{hilger2026cfear,
  title={{CFEAR-Teach-and-Repeat: Fast and Accurate Radar-only Localization}},
  author={Hilger, Maximilian and Adolfsson, Daniel and Becker, Ralf and Andreasson, Henrik and Lilienthal, Achim J},
  booktitle={2026 IEEE International Conference on Robotics and Automation (ICRA)},
  year={2026}
}

@INPROCEEDINGS{cen2018precise,
author = {Cen, Sarah and Newman, Paul},
year = {2018},
month = {05},
booktitle = icra,
pages = {1-8},
title = {{Precise Ego-Motion Estimation with Millimeter-Wave Radar Under Diverse and Challenging Conditions}},
doi = {10.1109/ICRA.2018.8460687}
}

@inproceedings{Checchin2009,
  title={{Radar Scan Matching SLAM Using the Fourier-Mellin Transform}},
  author={Checchin, Paul and G{\'e}rossier, Franck and Blanc, Christophe and Chapuis, Roland and Trassoudaine, Laurent},
  booktitle={Field and Service Robotics: Results of the 7th International Conference},
  pages={151--161},
  year={2010},
  organization={Springer}
}


\end{document}